\documentclass{article}
\newcommand{\documentdate}{11 V 2023}

\usepackage{a4wide,latexsym,graphicx,amsmath}
\usepackage{varioref}
\usepackage{ulem}
\usepackage{amsfonts}
\usepackage{xcolor}
\usepackage{float}
\title{A Block-Coordinate Approach of Multi-level Optimization
  with an Application to Physics-Informed Neural Networks}

\author{Serge Gratton\footnotemark[1],
        Valentin Mercier\footnotemark[2],
        Elisa Riccietti\footnotemark[3],
        Philippe L. Toint\footnotemark[4]}

\newcommand{\beqn}[1]{\begin{equation}\label{#1}}
\newcommand{\eeqn}{\end{equation}}
\newcommand{\req}[1]{(\ref{#1})}
\newcommand{\ms}{\;\;\;\;}
\newcommand{\tim}[1]{\;\; \mbox{#1} \;\;}
\newcommand{\pagenumbers}{\arabic{page}}
\renewcommand{\thepage}{\pagenumbers}
\newcounter{algo}[section]
\renewcommand{\thealgo}{\thesection.\arabic{algo}}

\newcommand{\algo}[3]{\refstepcounter{algo}
\begin{center}\begin{figure}[htbp]
\framebox[\textwidth]{
\parbox{0.95\textwidth} {\vspace{\topsep}
{\bf Algorithm \thealgo : #2}\label{#1}\\
\vspace*{-\topsep} \mbox{ }\\
{#3} \vspace{\topsep} }}
\end{figure}\end{center}}

\newcommand{\epr}{\hfill $\Box$ \vspace*{1em}}
\newcommand{\lthm}[2]{\vspace{\baselineskip} 
\noindent\framebox[\textwidth]{\parbox{0.95\textwidth}{
\begin{theorem} \label{#1} \rm #2 \end{theorem} } } \vspace{\baselineskip} }
\newcommand{\ii}[1]{\{ 1, \ldots, #1 \}}

\newcommand{\iibe}[2]{\{ #1, \ldots, #2 \}}
\newcommand{\hatf}{\widehat{f}} 
\newcommand{\barg}{\bar{g}} 
\newcommand{\calA}{{\cal A}} 
\newcommand{\calB}{{\cal B}} 
 
\newcommand{\calF}{{\cal F}}

\newcommand{\calO}{{\cal O}} 
 
\newcommand{\calS}{{\cal S}}

\renewcommand{\Re}{\hbox{I\hskip -2pt R}}
\newcommand{\smallRe}{\hbox{\footnotesize I\hskip -2pt R}}

\newcommand{\sfrac}[2]{{\scriptstyle \frac{#1}{#2}}}
\newcommand{\half}{\sfrac{1}{2}}

\newcommand{\al}[1]{{\footnotesize{\sf #1}}}
\newcommand{\tal}[1]{{\normalsize {\sf #1}}}

\newtheorem{theorem}{Theorem}[section]

\newcommand{\NN}{{\tt NN}}

\newcommand{\comment}[1]{}

\definecolor{myblack}{RGB}{53, 53, 53}
\definecolor{myblue}{RGB}{40, 75, 99}
\definecolor{myred}{RGB}{192, 50, 33}
\definecolor{myyellow}{RGB}{255, 166, 48}
\definecolor{mywhite}{RGB}{240, 237, 238}
\definecolor{mygreen}{RGB}{0, 102, 0}

\definecolor{green1}{RGB}{9, 82, 86}
\definecolor{green2}{RGB}{8, 127, 140}
\definecolor{green3}{RGB}{6, 167, 125}
\definecolor{green4}{RGB}{79, 109, 122}
\definecolor{green5}{RGB}{192, 214, 223}
\definecolor{violet}{RGB}{26,69,131}

\definecolor{checkgreen}{rgb}{0,0.6,0}
\definecolor{phase1}{rgb}{0.008,0.655,1.000}
\definecolor{phase2}{rgb}{0.016,0.75,0.700}
\definecolor{phase3}{rgb}{0.929,0.35,0.700}
\definecolor{icsyellow}{cmyk}{0.00,0.11,0.53,0.00}

\date{\documentdate}

\begin{document}

\maketitle

\renewcommand{\thefootnote}{\fnsymbol{footnote}}
\footnotetext[1]{Universit\'e de Toulouse, INP-ENSEEIHT, IRIT, Toulouse,
  France. Work partially supported by 3IA Artificial and Natural
  Intelligence Toulouse Institute (ANITI), French ``Investing for the
  Future - PIA3'' program under the Grant agreement ANR-19-PI3A-0004.
  Email: serge.gratton@enseeiht.fr.}
\footnotetext[2]{Universit\'e de Toulouse, ANITI, CERFACS, IRIT, Toulouse,
  and BRLi, France. 
  Email: valentin.mercier.gratton@enseeiht.fr.}
\footnotetext[3]{Universit\'e de Lyon, INRIA, ENSL, UCBL, CNRS, LIP
  UMR 5668, Lyon, France.
  Email: elisa.riccietti@ens-lyon.fr.}
\footnotetext[4]{Namur Center for Complex Systems (naXys),
  University of Namur, Namur, Belgium.
  Email: philippe.toint@unamur.be.}

\begin{abstract}
Multi-level methods are widely used for the solution of large-scale
problems, because of their computational advantages and exploitation
of the complementarity between the involved sub-problems. After a
re-interpretation of multi-level methods from a block-coordinate point
of view, we propose a multi-level algorithm for the solution of
nonlinear optimization problems and analyze its evaluation
complexity. We apply it to the solution of partial differential
equations using physics-informed neural networks (PINNs) and consider 
two different types of neural architectures, a generic feedforward network 
and a frequency-aware network. We show that our approach is 
particularly effective if coupled with these specialized architectures 
and that this coupling results in better solutions and significant 
computational savings.
\end{abstract}


\section{Introduction}

Many numerical optimization problems of interest today are large
dimensional, and techniques to solve them efficiently are thus an
active field of research. A very powerful class of algorithms for the
solution of large problems is that of multi-level methods. Originally,
the concept of a method exploiting multiple levels, i.e., multiple
resolutions of an underlying problem, was introduced for the solution
of large scale systems arising from the discretization of partial
differential equations (PDEs). In this context these methods are known
as multigrid (MG) methods for the linear case or full approximation
schemes (FAS) for the nonlinear one
\cite{BrigHensMcCo00,TrotOsteSchul01}. These schemes were later
extended to nonlinear optimization problems, in which context they are
known as multi-level optimization techniques
\cite{Nash00,GratSartToin05b,GratSartToin05c,GratSartToin08,CalaGratRiccVass21}.  The
central idea of all these approaches is to use the structure of the
problem in order to significantly reduce the computational cost
compared to standard approaches applied to the full unstructured
problem.

In this paper we introduce a new interpretation of multi-level methods
as block coordinate descent (BCD) methods: iterations at coarse levels
(i.e.,\ low resolution) can be interpreted as the (possibly
approximate) solution of a subproblem involving a set of variables
smaller than that required to describe the fine level (high
resolution).  We propose a framework that allows us to encompass
multi-level methods for several classes of problems as well as a unifying complexity
analysis based on a generic block coordinate descent, which is simple
yet comprehensive.

To illustrate the effectiveness of the proposed approach, we apply our
framework in the context of deep learning.  The idea of exploiting
multiple scales in learning has been explored for different kind of
networks. For instance,
\cite{KopaKrau22,HabeRuthHoltJun18,Wuetal20,vonPKopaKrau21} propose
multilevel methods for the training of deep residual networks
(ResNets), in which the multilevel hierarchy and the transfer
operators are constructed by exploiting a dynamical system's
interpretation. Multi-scale methods for convolutional neural networks
and recurrent networks have also been proposed in
\cite{HabeRuth17,TsunMairYu17} and \cite{ChunAhnBeng16},
respectively. Our focus in this paper is on physics-informed neural
networks (PINNs). These networks have been introduced in
\cite{RaisPerdKarn19} and have exhibited good
performance in practice, soon supported by theoretical results
\cite{MishMoli22,Shin20}. See
\cite{CaiMaoWangYinKarn21} for a comprehensive review on the topic.

Despite their success, the training of such networks may remain
difficult, in particular for highly nonlinear or multi-frequency
problems \cite{WangWangPerd21}.  In particular, choosing an
efficient detailed network architecture and an associated training
procedure is far from obvious, especially if the problem's solution
involves high frequency components (\cite{Xu20} has described
under the name of \textit{F-principle} why it might be so, see also
\cite{Rahaetal19,BasrJacoKasttKrit19,WangYuPerd20,FarhKazaWang22}).
While specific "frequency aware" network architectures, such as WWP
\cite{WangWangPerd21} structures and Mscale networks
\cite{LiuCaiXu20,Li20} have been proposed to
circumvent this latter difficulty, the mere size of the networks
necessary to represent solutions of PDEs with sufficient accuracy
still make their computationally efficient training very
challenging. Our objective is to make this
challenge more tractable.

\paragraph*{Contributions.}
In this context, the specific contributions of this paper may be summarized as follows.
\begin{enumerate}
\item We present a unifying framework for a large class of multi-level problems (Section~\ref{theory}).
\item We then introduce a suitable block-coordinate descent algorithm,
  and analyze its evaluation complexity bound under standard
  assumptions (Section~\ref{sec:BCD}).
\item We next investigate how this approach can be applied to PINNs
  for the solution of Laplace problems on complex geometries. 
\begin{enumerate}
    \item In a first step, we show that a multi-level technique based
      on alternate training of "coarse" and "fine" networks may bring
      substantial computational benefits (Section~\ref{sec:PINNs}).
    \item We then exploit frequency aware network architectures in
      this multi-level context, allowing the efficient solution of
      more complex problems (Section~\ref{MPINNV2}).
\end{enumerate}
Compared with standard (single-level) training, both approaches are
shown to yield better solutions at a much reduced computational cost. 
\end{enumerate}
Section~\ref{sec:conclusion} finally presents some conclusions and perspectives.

\section{A block-coordinate perspective on Galerkin multilevel optimization}
\label{theory}

Our purpose is to present a new (at least as far as we know)
but yet simple perspective on multilevel optimization using Galerkin approximations.
Given some space $\calF$ of continuous functions from $\Re^p$ to
$\Re^q$ and some objective function from $\calF$ into $\Re$,
our global aim is to compute a function $y$, which is a (possibly approximate)
solution of the variational problem 
\beqn{problem}
\min_{y\in\calF} f(y).
\eeqn
The objective function $f$ is often given by some norm of the residual of
a problem of interest (PDE, ODE, boundary value problem, linear system
or other) but other cases are possible (such as minimum surface or
contact problems, for instance). We assume that $\calF$ consists of
functions constructed by linearly or nonlinearly combining
elemental/basis functions using a parametrization involving the
parameters $x\in\Re^n$, so that $y$ is denoted by $y(x)$, whereas the
value of $y(x)$ at $z$ will be denoted by $y(x)(z)$.
The problem then reduces to  finding the value(s) of $x$ such that
$f(y(x))$ is  minimized.  In this paper, we focus on a class of
``splitting'' techniques whose objective is to reduce the
computational cost associated with this minimization.  More
specifically, we consider the case  where $y$ is viewed as the sum of
two terms $y_1$ and $y_2$,  themselves depending on their own sets of
parameters $x_1\in\Re^{n_1}$ and $x_2\in \Re^{n_2}$, that is
\beqn{split2}
y(x) = y_1(x_1)+y_2(x_2).
\eeqn
This yields the optimization problem
\[
\min_{(x_1,x_2)\in \smallRe^n} f(y_1(x_1)+y_2(x_2)),
\]
where $n = n_1+n_2$. We also associate with the splitting
\req{split2} the following ``approximation sets''
\beqn{Afull}
\calA_{12} = \big\{ y \in \calF \mid y(x)=y_1(x_1)+y_2(x_2)  \tim{for some}
  (x_1, x_2) \in \Re^n \big\}
\eeqn
as well as
\beqn{Ai}
\calA_i = \big\{ y \in \calF \mid y(x)=y_i(x_i) \tim{for some} x_i\in\Re^{n_i} \big\}
\ms (i=1,2).
\eeqn

While our present development is based on an additive structure of
the function $y$, other formulations could obviously be
of interest.  We limited the number of terms to two in order to
simplify exposition, but this not restrictive (as we discuss below). 

We now discuss some examples of this approach, in which we distinguish two main contexts.

\begin{description}
\item[The hierarchical context:]  
The terms $y_1$ and $y_2$ are constructed such that
  \begin{equation}\label{inclusion}
  \calA_2 \subset \calA_1 = \calA_{12}.
  \end{equation}
  This may occur in variety of cases, the simplest being the classical multigrid 
  framework in which one assumes that
  $f$ is a strictly convex quadratic and the $y_i$ are linear.  The quadratic's
  minimization is then equivalent to the solution of a positive definite
  linear system. The $y_i$ are constructed as linear combination of 
  basis functions $\{b_j\}_{i=1}^m$ of $\calF$ (typically from a finite-differences 
  or finite-elements basis), that is 
  \beqn{y1b}
  y_1(x_1) = \sum_{j=1}^m (x_1)_j\, b_j
  \tim{ and } 
  y_2(x_2) = \sum_{j=1}^m (Px_{2})_j\, b_j,
  \eeqn
  where $m=n_1$ is the dimension of $\calF$ and
  where $P$ is a $(n_1\times n_2)$ linear ``prolongation'' operator from a
  ``coarse'' space of dimension $n_2\leq n_1$ to the ``fine'' space of
  dimension $n_1$. In the multigrid framework, these  coarse and fine
  spaces often correspond to coarse and fine discretizations of an
  underlying continuous problem, but other interpretations such as
  domain decomposition, are possible. The quadratic optimization
  problem then becomes 
  \beqn{MG-full}
  \min_{(x_1,x_2)\in\smallRe^{n_1+n_2}} \half (x_1+Px_2)^TA(x_1+Px_2) + b^T(x_1+Px_2)
  \eeqn
  for some positive definite matrix $A$ and right-hand side $b$ depending of the basis
  $\calB = \{b_j\}_{i=1}^m$. One easily verifies that ``restricting'' the
  minimization to the $x_2$ variables  amounts to solving the
  $n_2$-dimensional problem
  \beqn{MG-coarse}
  \min_{x_2\in\smallRe^{n_2}}\half x_2^T PAP^T x_2 + (b+P^TAx_1)^Tx_2,
  \eeqn
  which is the usual Galerkin approximation of $f$ at the coarse
  level. We clearly have that
  \beqn{A1A2}
  \calA_2 = \left\{ \sum_{j=1}^m (Px_{2})_j \, b_j \right\}
  \subset  
  \left\{ \sum_{j=1}^m(x_{1,j}+(Px_{2})_j)\, b_j \right\}
  = {\rm span}(\calB)
 = \calA_1
 = \calA_{12},
 \eeqn
 ensuring (\ref{inclusion}).
  Classical multigrid methods then alternate approximate resolutions of the
  $n_2$-dimensional ``coarse level'' problem \req{MG-coarse} and the
  $n$-dimensional ``fine level'' one given by \req{MG-full}.

  A second, more nonlinear, case is when $f$ may no longer be
  a convex quadratic, but a smooth nonlinear function (which we assume
  is bounded below for consistency), while keeping \req{y1b} and its interpretation in terms of ``coarse'' and
  ``fine'' spaces. Reusing \req{y1b} and \req{A1A2} we
  may now consider  
  \begin{equation}\label{semil}
  \hatf(x_1,x_2)
  \stackrel{\rm def}{=} f\left(\sum_{j=1}^m x_{1,j} b_j+\sum_{j=1}^m (Px_{2})_j \, b_j\right)
  = f\left(\sum_{j=1}^m (x_{1,j}+(Px_{2})_j)\, b_j \right) 
  \stackrel{\rm def}{=} F(x_1+Px_2),
  \end{equation}  
  which is a reformulation of the original objective function 
  incorporating the dependence of the basis $\calB$ and, as above,
  alternatively perform iterations on the problems
  \[
  \min_{(x_1,x_2)\in\smallRe^{n_1+n_2}} \hatf(x_1,x_2)
  \tim{ and }
  \min_{
  \begin{array}{l}
    x_2\in\smallRe^{n_2}   \\
       x_1 \mbox{ fixed} 
  \end{array}
  } \hatf(x_1,x_2)
  \]
  and note that, because of \req{A1A2}, the first of these problems
  is equivalent (in the sense that they yield the same value for $y_1(x_1)+y_2(x_2)$) 
  to the lower-dimensional 
  \[
  \min_{
  \begin{array}{l}
    x_1\in\smallRe^{n_1}   \\
       x_2 \mbox{ fixed} 
  \end{array}
  } \hatf(x_1,x_2).
  \]
  This second approach has been explored in the framework of nonlinear
  optimization by the MG-OPT \cite{Nash00} and RMTR \cite{GratSartToin08} 
  methods. Both these algorithms consider a subproblem where a coarse-level model 
  of the objective function is approximately minimized. Our hierarchical 
  approach corresponds to the choice of the Galerkin Taylor models defined
  (for first and second order) by
  \[
  h_1(\delta x_2) = \sigma ( P^T \nabla_x F(x_1+P x_2))^T\delta x_2
  \]
  and
  \[
  h_2(\delta x_2) = \sigma (P^T \nabla_x F(x_1 + P x_2))^T\delta x_2
             +\half \sigma^2 \delta
             x_2^TP\nabla_x^2 F(x_1 + Px_2)P^T\delta x_2,
  \]
  where $\sigma$ is a fixed positive constant and $\delta
  x_2$ is an increment in $x_2$ from the point $(x_1,x_2)$. 
  Note that the form of $h_2$ is identical to that of \req{MG-coarse}.

 The framework just described is also closely related to the FAS
 approach \cite[p.~98]{BrigHensMcCo00} for solving a set of nonlinear
 equations. If we consider the equation $\nabla F (x) = 0$, the
 standard FAS approach would consist in solving at a given fine
 iterate $x_1$ the problem $P^T \nabla F (P^Tx_1  + x_2)P = \operatorname{rhs}$,
 where the right-hand side $\operatorname{rhs}$ is such that if $x_1$ solves the
 problem (i.e., annihilates the gradient of $F$), $x_2$ is  zero. This
 approach is different from ours in that the correction $Px_2$ in the
 coarse problem is added to $PP^T x_1$ in the coarse problem
 definition, and not to $x_1$ itself. Taking this into account,  we
 may consider the coarse equation in $x_2$,  $P^T \nabla F (x_1 +Px_2)
 = 0 $, for which  we obtain a formulation that is now in line with
 our hierarchical context since the derivative of this coarse equation
 are identical to those involved in the above definition of
 $h_2$. Note that the right-hand side of this equation is now zero
 since  $x_2=0$ solves the coarse problem when $\nabla F (x_1) = 0$. 

 In what follows, we will be especially interested in the case
  where \beqn{yi-NN} y_i(x_i) = \NN_i(x_i) \eeqn where $\NN_i(x_i)$ is
  a neural network of input $z$, parameters (weights and biases) given
  by $x_i\in \Re^{n_i}$, and output $\NN_i(x_i)(z)$. Neural networks are
  clearly nonlinear and nonconvex functions of the parameters $x_i$
  and the hierarchical context occurs when $y_2$ is a subnetwork of
  $y_1$. 
\item[The distributed context:] 
We may now abandon (\ref{inclusion}) and consider a situation where neither $\calA_1$ or $\calA_2$ is 
identical to $\calA_{12}$.  This is for instance the case when
(\ref{yi-NN}) holds, but $y_2$ is not a subnetwork of $y_1$ 
  and $n_1$ and $n_2$ (the number of
  network parameters in $y_1$ and $y_2$, respectively) are now
  independent of $m$. Our proposal is to use a similar methodology for
  this more complex case and alternate approximate minimizations on
  the ``$\calA_2$ subproblem'' given by
  \beqn{NN-A2} 
  \min_{
  \begin{array}{l}
    x_2\in\smallRe^{n_1}   \\
       x_1 \mbox{ fixed} 
  \end{array}
  } f(y_1(x_1)+y_2(x_2))
  \eeqn
  with that on the ``$\calA_1$ subproblem'' given by
  \beqn{NN-A1}
   \min_{
  \begin{array}{l}
    x_1\in\smallRe^{n_1}   \\
       x_2 \mbox{ fixed} 
  \end{array}
  } f(y_1(x_1)+y_2(x_2)).
  \eeqn
\end{description}

In all cases described above, the computational cost
of the overall minimization is expected to decrease because we may
choose $n_2$ (and, for the distrubuted case, $n_1$) to be
significantly smaller than $n$. Alternating standard minimization
steps on the fine $\calA_1$ level with cheap ones at the coarse
$\calA_2$ level is therefore computationally attractive,
\textit{provided the coarse steps significantly contribute to the
  overall minimization.}

We conclude this section by noting that it is obviously possible to
consider more general additive splittings of the form
\[
y(x,z) = \sum_{i=1}^s y_i(x_i,z)
\]
in our above developments.  In the the hierarchical
context, that is in the classical multigrid approach and in the
RMTR/MG-OPT algorithms, this is achieved by considering a hierarchy of 
nested approximations sets
\[
\calA_{\iibe{\ell}{s}} = \left\{ y \mid y
=\sum_{j=1}^m (x_{\ell,j}+P_1(x_{\ell+1,j}+P_2(\ldots
P_k(x_{s-1,j}+P_sx_{s,j})\ldots)))b_j\right\}
\]
for $\ell\in \ii{s}$, so that
$\calA_{\iibe{\ell+1}{s}}\subset\calA_{\iibe{\ell}{s}}$ for each $\ell\in\iibe{1}{s-1}$.

In the distributed context, one could consider the sets
\beqn{AFNs}
\calA_\calS = \left\{ y \mid y
= \sum_{i\in\calS} y_i(x_i) \right\}
\eeqn
for all nonempty subsets $\calS$ of $\ii{s}$.
This allows for a wide variety of set architectures such as the
``recursive'' one using $\calA_{\iibe{\ell}{s}}$ for $\ell \in \ii{s}$,
the ``flat'' one using $\calA_{\{\ell\}}$ for
$\ell \in \ii{s}$, or any mixture of these. In full generality, the
approximation sets \req{AFNs} need not be disjoint and it is then
useful, for a computationally
effective architecture, to identify which subsets
$\calS$ generate identical $\calA_\calS$ and to ignore those for which
$\sum_{i\in\calS}n_i$ (the dimension of the associated minimization
problem) is not minimal.

As a consequence of the above discussion, we see that a wide variety
of multilevel optimization  methods may be viewed as block-coordinate
minimization problems, where the blocks of variables are given by the
$x_i$.

\section{A generic BCD algorithm and its convergence}\label{sec:BCD}

We now examine why splitting minimization into alternate block
minimization can be useful, and consider the associated convergence
guarantees.  We first note that such guarantees are already available,
for linear multigrid  case
\cite{BrigHensMcCo00,Hack95,TrotOsteSchul01}  for  FAS-like methods
\cite{BorzKuni06}, as well as for MG-OPT \cite{Nash00} and RMTR
\cite{GratSartToin08,GratSartToin05c,GrosKrau11,CalaGratRiccVass21}.  The
objective of this section is to motivate and state a simple yet
comprehensive complexity analysis, covering the two contexts
described above.

The success of existing multilevel methods is based on exploiting a
``complementarity'' between the various minimization problems
involved, which we pursue as follows.
Given the overall problem \req{problem}, one starts by considering a
particular minimization method and isolate 
a class of problems for which this method is efficient.  In the
hierarchical context, this is typically the Gauss-Seidel or 
Jacobi method and the class of problems where such ``smoothing
methods'' are efficient is that of problems involving high-frequency
behaviour in the sought $y$ function of the underlying variable $z$
(see \cite[Chapter~2]{BrigHensMcCo00}). 
This suggests that one might wish to
split the problem (if at all possible) depending of its frequency
content.  To achieve this, one chooses (at least implicitly) the basis
$\calB$ (which spans $\calA_{12}$ and $\calA_1$) to be a Fourier basis
of $\calF$ and split the problem into finding the coefficients of the
high-frequency basis functions (using a smoothing method in $\calA_{12}=\calA_1$) and finding those of the
low frequency ones ($\calA_2$).  The key of the approach is to
transform the low-frequency subproblem into a high frequency one by
shifting frequencies, making the smoothing algorithm efficient also
for this subproblem.  This shift is usually achieved by considering a
coarser discretization of the underlying continuous problem (the
``coarse space''). Classical multigrid methods (and also nonlinear methods in
the hierarchical context) then alternate minimizations steps for the
high-frequency (``fine'') subproblem with minimizations in the
low-frequency (``coarse'') one, $\calA_2$.  Note that, since the
frequency shift may be obtained only by considering the underlying
geometrical space, an explicit expression in the Fourier basis in
unnecessary. Access to the high frequency basis
elements may be unavailable as such, but is included in
the contribution of the complete basis spanning $\calA_{12}$ and $\calA_1$. 

We propose to follow the same approach for the distributed context.
We then select a particular minimization method.  In our focus example
where $y(x)$ is a neural net, first-order training methods such as
variants of gradient descent are a natural
choice. Remarkably, it has been shown in \cite{Xu20} that such
methods are significantly more efficient for the solution of problems
whose solution involves \textit{low} frequencies. This observation,
called the ``F-principle'', is interesting on two accounts.  The first
is that it stresses the fact that frequency content is also
significant for neural net training, and the second is
that it acts ``in the opposite direction'' when compared to a
multigrid approach: low frequencies are favourable instead of being
problematic. One is then led to consider using a Fourier basis also in
the new context, split the problem into a part containing the
high-frequency basis (finding a suitable network $y_1(x_1)\in
\calA_1$) and its low frequency part (finding a network $y_2(x_1)\in
\calA_2$). and then to shift the frequencies of the subproblems
to make them more efficiently solvable.  As we will discuss
when presenting our numerical examples in Section~\ref{MPINNV2}, 
this can be achieved by  using Mscale networks \cite{LiuCaiXu20} 
and WWP \cite{WangWangPerd21}. As above, this (fortunately) 
does not require the explicit problem formulation in the Fourier basis, 
although one needs to be somewhat specific regarding the subproblems' frequency content.

We also note that, if the objective function $f$ were separable in
$x_1$ and $x_2$ in the selected basis $\calB$ (the Fourier basis, in
our examples), then only one of each subproblem minimization would be
sufficient for solving the overall problem. This is not
the case in general, but an argument based on quadratic approximation shows that
a weak coupling within $f$ between $x_1$ and $x_2$ improves the speed
of convergence for the block-coordinate minimization.

At each stage of the minimization of $f$, we may therefore compute
one or more step(s) for a subproblem defined by selecting a subset of variables or,
equivalently, a set of $y_i$, to (approximately) minimize, in a
typical block-coordinate descent (BCD) approach.

Pure cycling between the relevant subproblems is clearly an
option, and is the strategy most often used in the multigrid case,
where V or W cycles are defined to organise the
cycling. Alternatively, we may opt for some sort of randomized cycling
(see \cite{RichTaka11,Nest12} for the convex case) or follow (as we choose to do below) the procedure
used in the RMTR algorithm for the hierarchical case and select a
subset of variables for which the expected (first-order) decrease in
the objective function (as measured by the norm of the objective's
gradient with respect to the variables in the subset) is sufficiently large.

We may therefore consider a simple block-coordinate descent algorithm for minimizing $f$, where we
use a second subscript for $x$ to denote iterations numbers, and where
we have limited the exposition to the bi-level/blocks case.

\algo{ML-BCD}{Multilevel Optimization (\tal{ML-BCD})}{
\begin{description}
\item[Step 0: Initialization: ] An initial point
$(x_{1,0},x_{2,0}) \in \Re^n$, a threshold $\tau\in(0,1)$ and a gradient accuracy threshold
 $\epsilon \in (0,1]$  are given.
Set $k=0$.
\item[ Step 1: Termination test. ] Evaluate the gradients
$g_k= (g_{1,k}^T,g_{2,k}^T)^T$ where $g_{i,k} = \nabla_{x_i}^1 f(y_1(x_{1,k}),y_2(x_{2,k}))$.
Terminate if $\|g_k\| \leq \epsilon$.
\item[Step 2: Select a subproblem and a subproblem termination rule. ]\mbox{}\\
For instance,\\
\hspace*{1mm}  select $i$ such that $\|g_{i,k}\|>\tau\|g_k\|$ and choose to minimize $f(x_1,x_2)$ as a function of
$x_i$.

Also select a termination rule for the chosen subproblem.
\item[Step 3: Approximately solve the chosen subproblem. ]
  Apply a monotone first-order minimization method to the chosen
  subproblem, starting from $(x_{1,k},x_{2,k})$
  and iterate for $p$ iterations until the selected termination rule
  is activated. This yields a new iterate
  $(x_{1,k+p},x_{2,k+p})$ such that $f(x_{1,k+p},x_{2,k+p})<f(x_{1,k},x_{2,k})$.
\item[Step 4: Loop. ] Increment $k$ by $p$ and go to Step~1.
\end{description}
}

In the form stated above, the \al{ML-BCD} algorithm requires computing
the full gradient at every major iteration (i.e.,\ iteration where Step~2 is used), at variance with a pure
(potentially randomized) cycling where only the successive subproblem's
gradients need to be computed.  Thus the number of major iterations
must remain small compared to the total number of iterations (as
indexed by $k$) for this approach to be useful.

Also note that the subproblem termination criterion may take different
forms: the number of subproblem iterations may be limited, a threshold
may be imposed on the norm of the subproblem's gradient and/or on the
objective function decrease, below which
the subproblem minimization is terminated, or any combination of
these. In any case, it does not make sense to continue the subproblem
minimization if the subproblem's gradient becomes smaller that
$\tau \epsilon$.

The convergence theory for block-coordinate optimization has a long
history, starting with a famous paper by Powell \cite{Powe73} showing
that the method may fail on nonconvex continuously differentiable functions.  While
the theory was further developed for the convex case (see the
excellent survey \cite{Wrig15} and the references therein), it
was only recently that further progress was made for nonconvex functions, overcoming Powell's
reservations, and that a worst-case complexity analysis (implying
convergence) was produced \cite{AmarAndrBirgMarc21}.  The idea is quite
simple and rest on the notion of ``sufficient descent'', which
requires, when using first-order methods, that 
\[
f(x_k)-f(x_{k+1}) \geq \kappa \|g_k\|^2,
\]
for all $k\geq 0$, where $\kappa$ is a positive constant only
depending on $f$ itself.  This sufficient descent is, in particular,
guaranteed by the Lipschitz continuity of $\nabla_x^1f$ along the path
of iterates $\cup_{k \ge 0} [x_k,x_{k+1}]$ (see \cite[Notes for Section~2.4]{CartGoulToin22}), in
which case $\kappa$ is proportional to the inverse of the gradient's
Lipschitz constant. For the sake of completeness, we give a simple proof 
in appendix for a version of the \al{ML-BCD} algorithm using fixed-stepsize 
gradient descent in Step~3 (i.e.\ when, for all $k \geq 0$,
\begin{equation}\label{x-recur}
x_{k+1}=x_k-\alpha \barg_{i,k}
\end{equation}
where $\barg_{i,k} = (g_{1,k}^T,0)^T$ if $i=1$ and $(0,g_{2,k}^T)^T$
if $i=2$). This proof rephrases a standard "single block" result (see for instance 
\cite[Example 1.2.3]{Nest04}) for the BCD case. The key complexity result 
can be stated as follows. 

\lthm{th:complexity}{Suppose that $f$ is continuously differentiable
  with Lipschitz continuous gradient on the path of iterates $\cup_{k
    \ge 0} [x_k,x_{k+1}]$ generated by the \al{ML-BCD} algorithm with
  fixed stepsize, and that it is bounded below by $f_{\rm low}$.  Suppose also that the
  stepsize $\alpha$ is small enough to ensure $\alpha < 1/L$, where
  $L$ is the gradient's Lipschitz constant and that the $i$-th
  subproblem is terminated at the latest as soon as $\|g_{i,k}\| \leq
  \epsilon/\sqrt{2}$, Then the \al{ML-BCD} algorithm requires at most
  $\kappa_* \epsilon^{-2}$ iterations to produce an iterate $x_k$ such
  that $\|\nabla_x^1f(x_k)\|\leq\epsilon$, where $\kappa_*$ is a
  positive constant only depending of $\alpha$, $\tau$ and the initial gap
  $f(x_0)-f_{\rm low}$.
}

Other versions
of the algorithm including more elaborate globalization techniques
such as linesearches, trust-regions or
adaptive regularization are also possible and yield the same
$\kappa_*\epsilon^{-2}$ complexity bound for  different values
of the constant $\kappa_*$ (we then say that their complexity is  $\calO(\epsilon^{-2})$).

The situation is more involved and the complexity bound worse (when
it exists) as soon as the function is not Lipschitz continuous on the
path of iterates, and is for instance discussed (for the single block case) 
in \cite{JordLinZamp22,TianManC21,ZhanLinJegeSraJadb20}. 
The recent paper \cite{KongLewi22} analyzes the
complexity of a ``first-order'' monotonic descent method applied to a wide
class of functions\footnote{Technically, the objective function must
be bounded below, have a bounded directional subgradient map and a
  finite nonconvexity modulus.} . The method assumes that one can evaluate, for any
point $x$ and any direction $d$, the value of $f(x)$, its directional
derivative along $d$ $f'(x,d)$ and a ``directional subgradient''
$G(x,d)$ such that its inner product with $d$ returns $f'(x,d)$.
Under these conditions, the method achieves (Goldstein \cite{Gold77}) $\epsilon$-approximate
optimality in at most $\calO(\epsilon^{-4})$ such evaluations. Unsurprisingly, the
convergence proof once more relies on showing that it is possible to
obtain ``sufficient descent'', in this case given by a multiple of $\epsilon$, albeit at
a possible cost of $\calO(\epsilon^{-3})$ evaluations. We refer the reader to
\cite{KongLewi22} for details. The monotonic nature of the algorithm
then implies (as is the case for the simple proof in appendix) that
sufficient descent obtained in the solution of the subproblem before termination
translates to sufficient descent on the complete problem, so that the
complexity result obtained by \cite{KongLewi22} for single block
minimization extends to the case where there is a (bounded) number of blocks.

Interestingly, Theorem~\ref{th:complexity} subsumes and simplifies the convergence 
theory for RMTR \cite{BrigHensMcCo00,GratSartToin08} by recasting this latter method (when used 
with first-order Galerkin low-level approximations) as a trust-region 
BCD algorithm in the complete space.
Also note that the use of Galerkin  approximations in this case avoids the need 
of a "tau correction" or "first-order coherency" condition \cite{BrigHensMcCo00,GratSartToin08} which 
is typically requested for less structured low-level approximations.

\section{Application to Physics-Informed Neural Networks}\label{sec:PINNs}

In recent years, using neural networks has emerged as an alternative
to classical methods for solving partial differential equations
(PDEs). In particular, the physics-informed neural networks (PINNs)
have raised significant interest \cite{RaisPerdKarn19}. 
They have the advantage of being able to exploit physical knowledge to solve equations without using simulation data.
This is why we choose to use this technique to illustrate our
multilevel approach.  Our presentation proceeds in two stages: after a
brief introduction to PINNs and  their multilevel version, we first
focus on showing the advantage of alternate training of "coarse" and
"fine" networks in the hierarchical context of Section~\ref{theory},
before showing how refined use of the frequency content (in a
distributed context) may yield further benefits. 

\subsection{Physics-informed neural networks}

Given a domain $\Omega\subset\mathbb{R}^d$, we consider the following differential system:
\begin{equation}
\label{general_problem}
    \left\{
    \begin{aligned}
    &\mathcal{L}(u(z))= r(z)  \text{ in } \Omega  \\
    &\mathcal{B}(u(z)) = g(z)  \text{ on }  \Gamma     
    \end{aligned}
  \right.
\end{equation}
where $\Gamma$ is the boundary of $\Omega$, $\mathcal{L}$ and
$\mathcal{B}$ are two (possibly nonlinear) differential operators and
$r$ and $g$ are two given functions.  

PINNs approximate the solution of the problem by a sufficiently smooth
neural network $y(x): \mathbb{R}^d \rightarrow \mathbb{R}$. The neural
network is trained by minimization of a loss that takes into account
the physical information contained in the PDE. Specifically, denoting
$Z_\Omega=\{z_\Omega^j\}_{j=1}^{N_{\Omega}}$,
$Z_{\Gamma}=\{z_{\Gamma}^j\}_{j=1}^{N_\Gamma}$ a set of training
points sampled in $\Omega$ and $\Gamma$ respectively, the loss
function is defined as 
\begin{equation}\label{loss_general}
\begin{aligned}
f(x) = \frac{\lambda_{\Omega}}{N_\Omega}\sum_{j=1}^{N_\Omega}[\mathcal{L}(y(x)(z_\Omega^j))-r(z_\Omega^j)]^2 +\frac{\lambda_{\Gamma}}{N_\Gamma}\sum_{j=1}^{N_\Gamma}[\mathcal{B}(y(x)(z_{\Gamma}^j))-g(z_{\Gamma}^j)]^2,
\end{aligned}
\end{equation}
where $\lambda_{\Omega}$, $\lambda_{\Gamma}$ are some positive weights
which balance the contribution of the residual of the PDE and the
residual of the boundary conditions. The differentiability properties
of the neural networks are exploited to compute explicitly the
differential operators $\mathcal{L}(y(x))$ and $\mathcal{B}(y(x))$,
which are then evaluated on the set of training points.   

Our objective is then to use several PINNs in conjunction with the
\al{ML-BCD} algorithm, broadly mimicking multigrid methods, in the
hope of obtaining similar computational advantages.  We define
multi-level PINNS (MPINNs)  as follows.  

Consider a feedforward neural networks with $N - 1$ hidden layers. 
Let $d_j \in \mathbb{N}$ be the number of hidden neurons in the $i$-th
hidden layer for $j = 1, . . . , N - 1$ and let $d_0$ and $d_N$ be the
number of neurons of the input and the output layers,
respectively. Let $W^j \in \mathbb{R}^{d_j\times d_{j-1}}$ be the
matrix of weights between the $(j - 1)$-th and the $j$-th layers for
$j =1,\dots, N$.  We denote the set of all such networks by $H^{N,\{d_j\}}$.  
Assuming a method is selected to minimize the loss functions
(\ref{loss_general}), we may now use the \al{ML-BCD} algorithm by
selecting our neural network as  
\begin{equation}\label{mpins-sum}
y(x)(z) = \sum_{i=1}^s y_i(x_i)(z)
\end{equation}
with $y_i \in H^{N_i,\{d_{j,i}\}}$. 

\subsection{Alternating training in a hierarchical context}
\label{mpinnsv1}

We start by considering the question of whether imitating the
multigrid approach of alternating between a coarse and a fine grid can
be computationally interesting.  

\subsubsection{Test problems}

Inspired by \cite{RapaSamt20}, we perform our experiments
on several instances of the Poisson problem with source term $r$ in
the domain $\Omega$ and Dirichlet/Neumann conditions on its boundary
$\Gamma$.  Moreover, we assume that $\Omega$ contains a closed
embedded boundary $\Gamma_i$ dividing $\Omega$ in $\Omega_e$ (exterior
of $\Omega_i$) and $\Omega_i$ (interior of $\Omega_i$), so that the
total boundary is given by $\Gamma = \Gamma_e \bigcup \Gamma_i$. The
problem is thus stated as 
\begin{equation}
\label{probleme_poisson}
  \left\{
  \begin{aligned}
      &\Delta u(z) = r(z)\text{ in }\Omega \\   
      &au(z) + b\frac{\partial u }{\partial n}(z) = g(z)\text{ on }\Gamma_e \\
      &cu(z) + d\frac{\partial u }{\partial n}(z) = h(z)\text{ on }\Gamma_i \\
  \end{aligned}
  \right.
\end{equation}
where $\frac{\partial u}{\partial n}= \nabla u^Tn$ with $n$ being the boundary normal vector pointing into $\Omega$.
The domain for the test problem is the square $\Omega = [-1,1]^2$ with
an embedded circle of radius $R=0.5$ centered at the origin defining
$\Omega_i$. We consider Dirichlet boundary condition on $\Gamma_e$ and
$\Gamma_i$.  We choose $r$ to ensure that exact solution is  
$$ u(z) = cos(\alpha\pi z_1+ \pi z_2) + cos(\pi z_1+\beta \pi z_2)$$
where $\alpha$ and $\beta$ are integers defining the frequency content of the solution. 

\subsubsection{Networks architectures}

In this first set of experiments, we simplify (\ref{mpins-sum}) to
$$y(x)(z) = y_1(x)(z)+y_2(x)(z)$$
where $y_1 \in H^{3,\{d_{j,1}\}}$ is the "fine" network and $y_2 \in H^{3,\{d_{j,2}\}}$ is the "coarse" one.
In accordance with our earlier discussion, we choose $y_2$ smaller
than $y_1$ in the sense that $y_2$ is an (independent) copy of a
subnetwork of $y_1$. Thus our setting is hierarchical (in the sense of
Section~\ref{theory}) and $\calA_2 \subset \calA_1=\calA_{12}$. 

We consider three different MPINNs corresponding to different sizes of
the coarse network, while keeping the size of the fine
network fixed. We compare them  to a standard PINN network of size
approximately equal to the total number of parameters in the fine and
coarse networks, as well as a network of the same size as the fine
network. 

\begin{table}[H]
\centering
\begin{tabular}{|c|c|c|c|}
\hline
Experiment's & \multicolumn{2}{c|}{Network(s)' size $(d_1,d_2,d_3)$} & Number of parameters\\
   name    &                      Fine & Coarse            & \\
\hline
ML1 & (140,140,140) & (140,140,140) & 40,040 + 40,040 \\
ML2 & (140,140,140) & (100,100,100) & 40,040 + 20,600 \\
ML3 & (140,140,140) & (70,70,70)    & 40,040 + 10,220 \\
SL1 & \multicolumn{2}{c|}{(140,140,140)} & 40,040 \\
SL2 & \multicolumn{2}{c|}{(200,200,200)} & 81,200 \\
\hline
\end{tabular}
\caption{Network architecture for the experiments with alternating training in the hierarchical context
\label{expe1}}
\end{table}

Table~\ref{expe1} details the five architectures tested in our experiments and the size of the involved networks:  
ML1, ML2 and ML3 are the multilevel ones (training two networks 
using the \al{ML-BCD} algorithm) and SL1 and SL2 are single level ones
(standard training of a single network) provided for comparison. All
hidden layers use the $\tanh$ activation function, thereby ensuring the
necessary differentiability properties. 

\subsubsection{Training setup}\label{setup}

Training points are sampled using the Latin hypercube sampling in
$\Omega$ and $\partial \Omega$. Here we have chosen  to use the same
grid to train the coarse and the fine networks with $N_\Omega = 50000$
points sampled in the domain and $N_\Gamma = 4000$ points sampled on
the boundary.  

At each epoch we use a random subset of these points composed of
$2000$ inner points and $500$ boundary points. We have chosen the
coefficients $\lambda_{\Omega}=\lambda_{\Gamma}=1$  to weight internal
and boundary losses. 

To evaluate the accuracy of the different models, we consider a set of
testing points $\{z^t\}_{t=1}^T$ with $T=30000$, randomly chosen using
the Latin hypercube sampling in $\Omega_e$, and consider the mean
squared error given by 
\[
MSE = \frac{1}{T}\sum^T_{t=1}(y(x)(z^t) - u(z^t))^2
\]

All networks are trained with Adam optimizer \cite{KingBa15}. The
initial learning rate is set at $2\times 10^{-4}$ with an exponential
decay of $0.99999$ at each epoch for all networks. 
All our codes are implemented in TensorFlow2 (version 2.2.0) and run
with a single NVIDIA GeForce GTX 1080 Ti. 

It is important to notice that using an alternate training strategy
requires special care when setting the hyperparameters of the
optimizer. Indeed, each time we select a sub-network to train, the
optimization problem also changes. To ensure a fair comparison with
standard network training procedures, we decided to maintain the
current learning rate when restarting the solver. This allowed us to
maintain the property that the learning rate gradually decreases over
time, which is a common technique used to prevent the model from
overshooting the optimal values and enhances convergence. By keeping
the decayed learning rate consistent across restarts, we were able to
compare the performance of the alternate training strategy with the
standard one in a consistent manner. For other parameters, we decide
to use Adam as a black box and to restart it each time we change
subproblem, that is we do not transfer the specific hyperparameters
that are specific to this optimization method, such as momentum. This
means that we have to initialize the optimizer from scratch with
default hyperparameters (except for the learning rate) for each
subproblem. The consequence of such a restart is a peak in the loss
at the first iteration of the optimization process, which is a common
behavior when initializing an optimizer with random or default
values. However it remains an open question to find the best strategy
to tune these hyperparameters: is it better to restart the optimizer
or to find a good way to transfer the parameters from one subproblem
to the other?  
We also estimate the number of floating-point operations performed as the number of FLOPs required for the matrix-vector product operations during the forward pass. For MPINNs, it takes into account both the operations
performed at fine and at coarse levels. For instance, for a network with two layers, $d_h$ neurons in the layers, an output size of 1, and $N$ training points of size $d$, the number of FLOPs for the forward pass would be computed as:
$$\text{FLOPs} = N \times (2d \times d_h + d_h^2 + d_h \times 1)$$
We select the subnetwork to train at each cycle according to Step 2 of
the \al{ML-BCD} algorithm.  Moreover, we chose to terminate the
subproblem training after a fixed number of epochs on each problem
(see numerical results). We also chose $\alpha=2$ and $\beta=4$. For
each experiment, we alternately performed 2000 epochs on $y_1$ and
2000 epochs on $y_2$. 

\subsubsection{Results}

\begin{figure}[H]
\begin{minipage}[b]{0.45\linewidth}
\centering
\includegraphics[width=\linewidth]{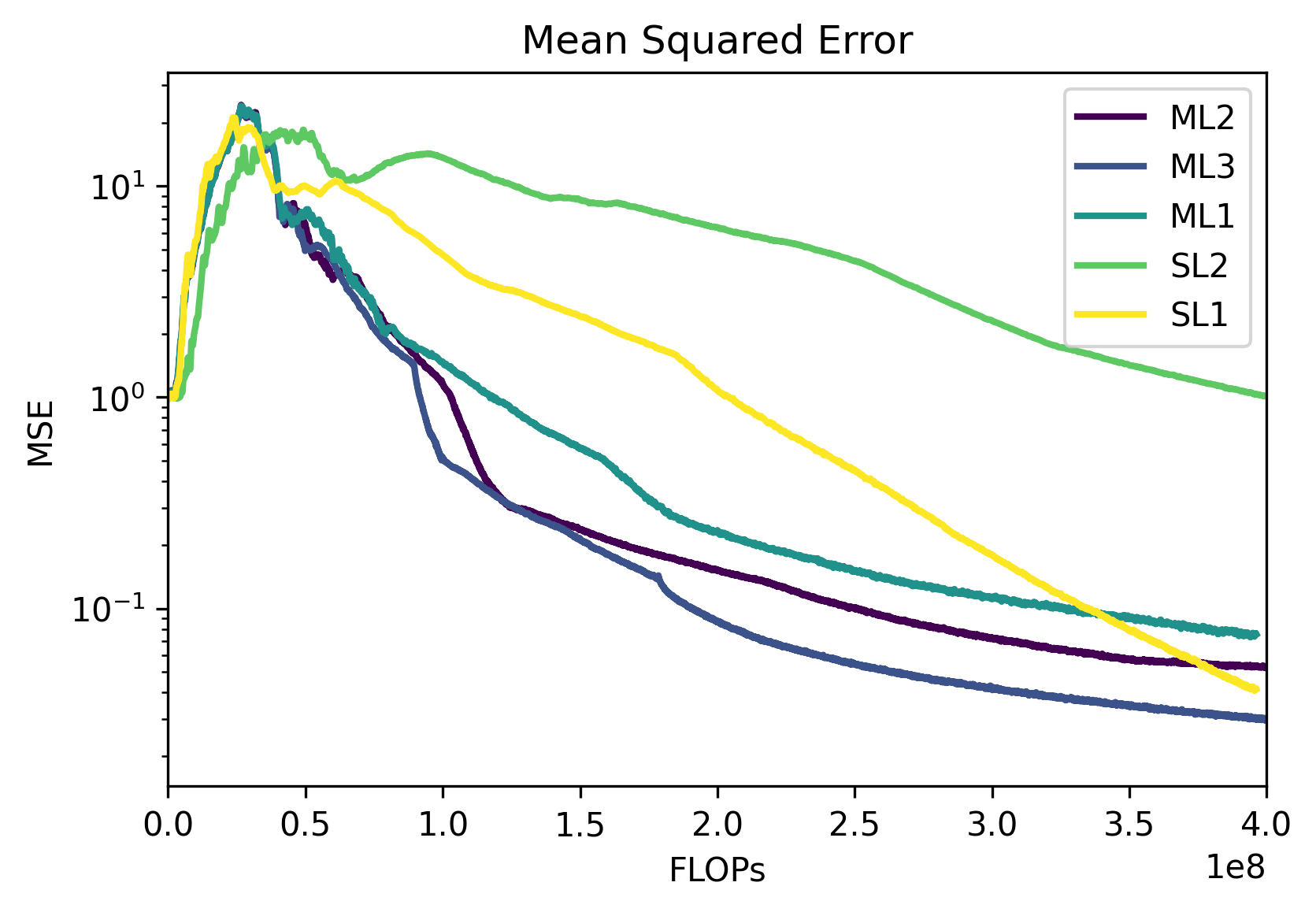}
 \caption{Evolution of the MSE as a function of the computational cost for our different models}
\label{fig:msempinnv1}
\end{minipage}
\hspace{0.5cm}
\begin{minipage}[b]{0.45\linewidth}
\centering
\includegraphics[width=\linewidth]{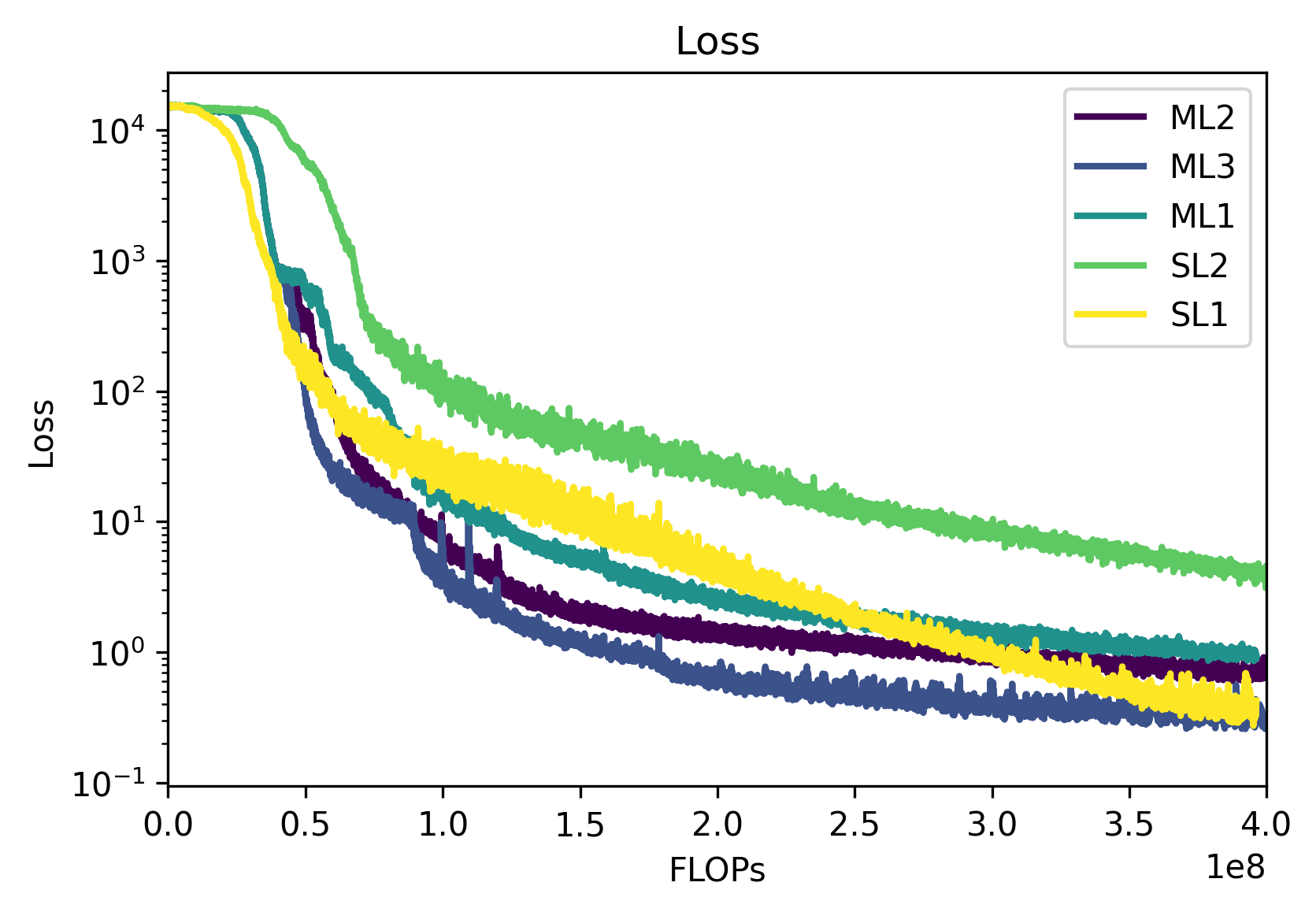}
\caption{Evolution of the loss as a function of the computational cost for our different models}
\label{fig:lossmpinnv1}
\end{minipage}
\end{figure}

The results of the test are reported in Figures  \ref{fig:msempinnv1}
and \ref{fig:lossmpinnv1}. We see that, at equivalent computational
cost, MPINNs (MLs) converge significantly faster than conventional
PINNs (SLs). It is worth noting that the PINN with fewer parameters
(SL1) converges faster than the larger one (SL2). The choice of the
coarse network's dimension also seems to affect the speed of
convergence, smaller coarse networks yielding better results in these
tests. 
\label{results1}

\begin{figure}[H]
    \centering
    \includegraphics[width=\linewidth]{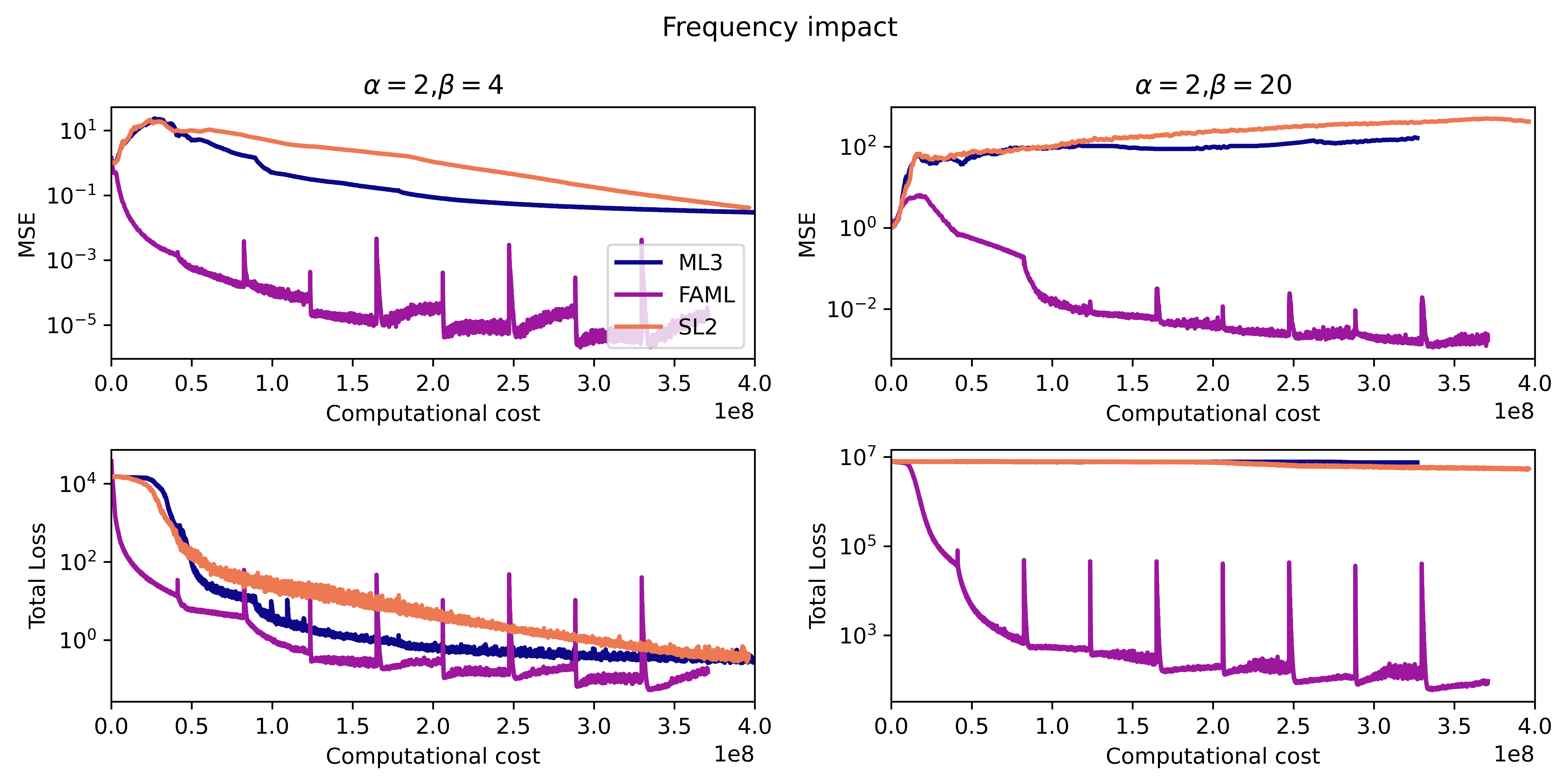}
    \caption{Results of our method on the Poisson problem
      (\ref{probleme_poisson}) for different frequencies. The method
      indicated as FAML (Frequency-Aware-Multilevel) is the method
      proposed in the forthcoming Section \ref{MPINNV2}.} 
    \label{fig:fprinciple}
\end{figure}

These result may seem encouraging, but this approach remains limited
in that it does not overcome a limitation inherent to classical neural
network training: high-frequency fitting. This is highlighted in
Figure \ref{fig:fprinciple} where we test our method on problem
(\ref{probleme_poisson}) with parameters $\alpha=2$ and $\beta=20$.  

An alternative approach is thus warranted, which we develop in the next section.

\section{A ``frequency-aware'' distributed approach}
\label{MPINNV2}


In the hierarchical cases detailed in Section~\ref{theory}, the reduction in computational cost is typically directly proportional to the separation of frequencies. For instance, in multigrid methods, switching to a coarse grid helps to target low frequencies that converge slowly in the fine problem. This does not seem to be the case for neural networks.  The F-Principle \cite{Xu20} states that "deep neural networks often fit target functions from low to high frequencies during the training process", which is problematic in some cases and seems to affect also MPINNs, leading to a slow convergence, as illustrated at the end of the previous section. The simple alternation of coarse and fine problems doesn't seem to be sufficient in the context of neural networks.

Several papers have addressed the issue related to the F-Principle by
proposing architectures that transform high frequencies of the problem
into low frequencies, thereby allowing a more efficient use of the
neural networks. Inspired by these papers, we propose a
frequency-aware architecture that retains the computational cost
savings of multilevel training while incorporating the frequency
separation aspect of classical methods. 

\subsection{A frequency-aware network architecture}

Our architecture proposal is mainly inspired by Multi-scale deep
neural networks (MscaleDNNs)  \cite{LiuCaiXu20} and WWP
\cite{WangWangPerd21}, which were designed to mitigate the
effect of the F-principle in the standard (single level) context.  
The basic idea is to transform high frequency learning in low
frequency learning and to facilitate the separation of the frequency
contents of the target function. As a result, these networks show
uniform convergence over multiple scales. 

The MscaleDNNs architecture achieves this objective thanks to two main ingredients:
\begin{itemize}
    \item radial scaling in the frequency domain: the first layer is
      separated into $N$ parts, each  receiving a differently scaled
      input. Several variants of MscaleDNNs have been proposed, we choose to focus
      here on the most efficient one, which uses parallel sub-networks,
      each dedicated to a specific input scaling. 
    \item the use of wavelet-inspired and frequency-located activation
      functions. These functions, with compact support, have good
      scale separation properties. 
\end{itemize}  

WWP networks were proposed by Wang,
 Wang, and Perdikaris. They use the same principle of input scaling, combined with soft Fourier mapping (SFM): 
\[
  \gamma (z) = s\; \begin{bmatrix}
           \cos(z) \\
           \sin(z) \\
         \end{bmatrix}
\]
with $s\in [0,1)$ a relaxation parameter. 

 A "Fourier features network" was first proposed in
 \cite{Tancetal20}, which uses a random Fourier features
 mapping $\gamma$ as a coordinate embedding of the inputs, followed by
 a conventional fully-connected neural network. This method did
 mitigate the pathology of spectral bias and helped networks learn
 high frequencies more effectively. Recent advancements have made it
 possible to model fully connected neural networks (and thus PINNs) as
 kernel regression. The authors of \cite{Tancetal20}
 suggested that using Fourier mapping affects the width of the kernels
 and thus the network's capacity to capture high frequencies. This
 idea was extended to PINNs in \cite{WangWangPerd21} with
 convincing results, provided that the fixed scalings are too far from the frequencies
 contained in the solution.

Inspired by their success, we propose here an architecture that
combines the positive features of both methods. Our architecture's
output is the sum of several networks that use different scaling
vectors, specializing each network for different frequency scales
matching the structure presented in our theory in Section \ref{theory}. 
To mitigate possible convergence problems arising from a bad 
choice of the fixed scalings, we have chosen to add learnable weights to the input
 scaling integers using SFM as a classic activation function.  

For low frequency resolution, we choose a classical network using hyperbolic tangent activation functions. For the other networks, we use SFM activation functions for the first layer associated with a scaling from a centred normal distribution whose variance grows with the targeted frequencies. The other layers of the networks use hyperbolic tangent activation functions, thus ensuring our differentiability requirements.  We refer to this architecture as Parallel-WWP (P-WWP).
 Unfortunately, the wavelet-based and frequency-based activation
 functions utilized in the MSCALE are not continuously differentiable,
 and thus fail to satisfy our theoretical assumptions. We have however
 included some (successful) experiments conducted using these
 functions in Appendix~\ref{mscalefaml}.  

A similar architecture, based on a lower number of subnetworks, has
been proposed in \cite{LiXuZhang21} for the deep Ritz method
\cite{Yuetal18}, which produces a variational solution to problem
\eqref{general_problem}. To the best of our knowledge, this is the
first time this architecture has been proposed for PINNs. 

\begin{figure}[H]
    \label{sdnn}
    \centering
    \includegraphics[width=\linewidth]{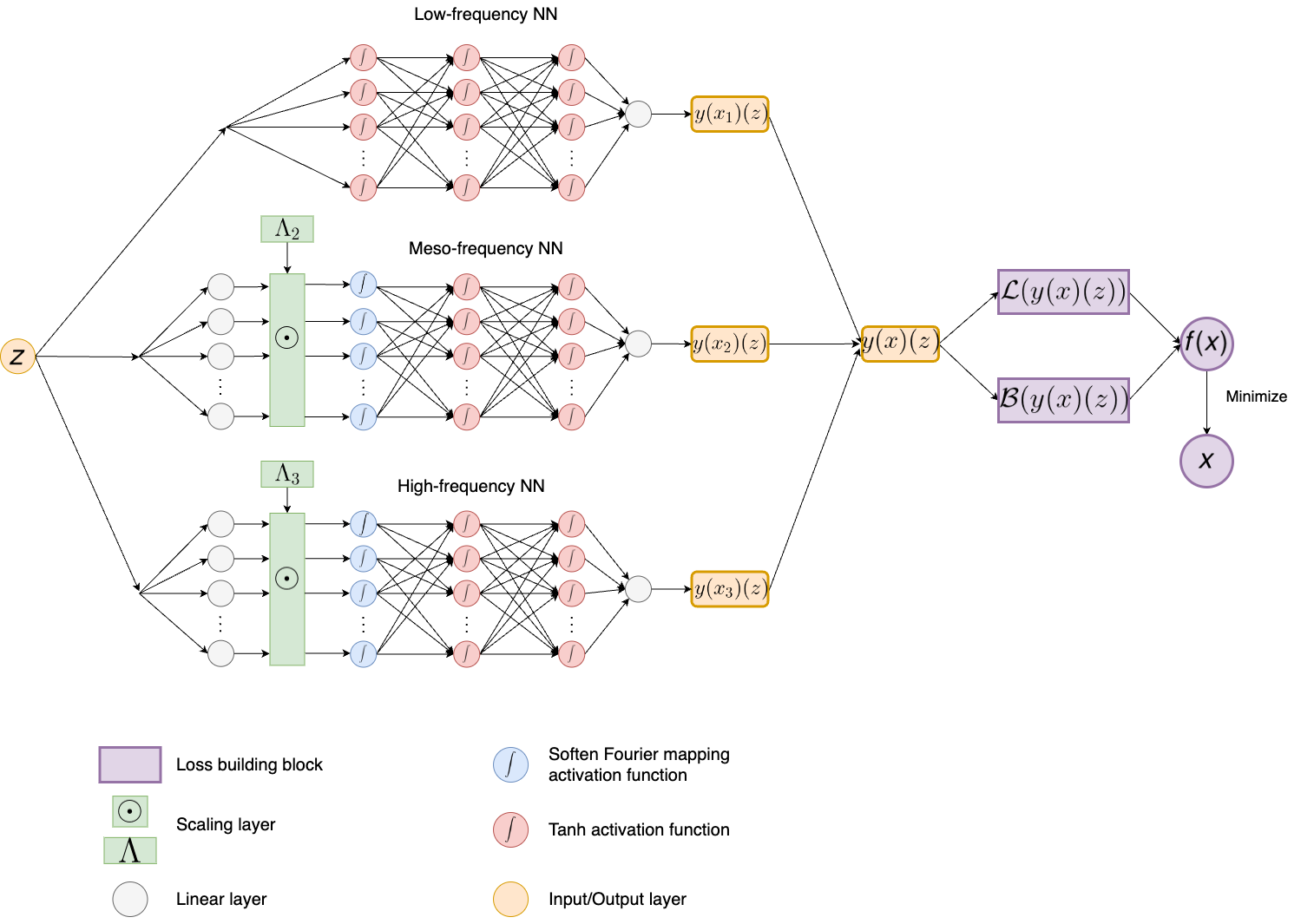}
    \caption{An example of P-WWP architecture }
    \label{fig:pwwp}
\end{figure}

An example of our architecture is illustrated in
Figure~\ref{fig:pwwp}, which consists of three sub-networks, targeting
three different frequency ranges defined by their input scaling
$\Lambda_i$. The lowest frequency network employs $\tanh$ activation
functions, while the others use $SFM$ and $\tanh$. 

To be effective, this architectures must cover most of the frequencies
contained in the a priori unknown PDE's solution. Therefore, a broad
range of frequencies must be covered, employing a large number of
neurons, as the target frequencies are defined by the input
scalings. As a result, the considerable improvement in accuracy
provided by parallel frequency-aware architectures is obtained at the
price of a significant increase in the training's computational cost,
making it an ideal candidate for our multilevel approach. 

These architectures can be easily incorporated into our framework,
where the global network is defined as a sum of $y_i$ sub-networks,
each responsible for a frequency range defined by its input
scaling. Because of this latter characteristics, the resulting
approach now belongs to the distributed context discussed in
Section~\ref{theory}.  
In particular, we refer to the combination of the P-WWP network with
the \al{ML-BCD} algorithm as the FAML (Frequency Aware Multi-Level)
method. We use the same selection and termination criteria as in
Section~\ref{mpinnsv1}. When several sub-networks are available, the
network with the highest ratio $\|g_{i,k}\|/\|g_k\|$ is chosen.  

\subsection{Numerical results}\label{sec:num}

We now illustrate the efficiency of the FAML method on some examples
of the form (\ref{probleme_poisson}) defined in
\cite{RapaSamt20}. We first describe the test problems
themselves, then specify the considered network architectures and the
training setup before describing and commenting the obtained results.   

\subsubsection{Test problem 1: Circle embedded in a square domain}

The domain for the first problem is the square $\Omega = [-1,1]^2$
with an embedded circle of radius $R=0.5$ centered at the origin
defining $\Omega_i$. We consider Neumann boundary conditions on
$\Gamma_i$ and Dirichlet boundary condition on $\Gamma_e$.  

\begin{figure}[H]
    \centering
    \includegraphics[width=0.8\linewidth]{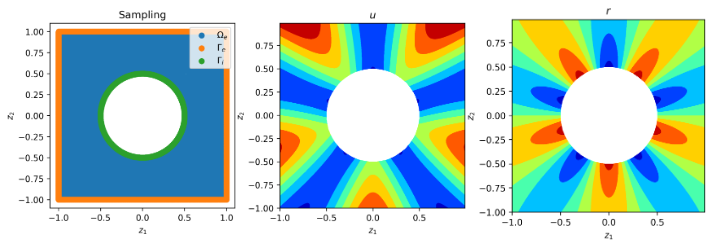}
    \caption{Test problem 1 (cf. \eqref{probleme_poisson}, \eqref{pb1}). Left: the domain $\Omega$ and its subdomains. Center: the solution $u$. Right: the source term $r$.}
    \label{fig:pb1}
\end{figure}

The source term and boundary term are given by 

\begin{equation}\label{pb1}
   \left\{
   \begin{aligned}
      &r(\rho,\omega) = D(k^2-n^2)\rho^{k-2}\sin(n\omega) \text{ in }\Omega \\   
      &g(\rho,\omega) =  -\frac{Dk}{n}\left(\frac{\rho}{R}\right)^{(n-k)}\rho^k\sin(n\omega)+Rq\log \rho\text{ on }\Gamma_e \\
      &h(\rho,\omega) = 1 \text{ on }\Gamma_i \\
   \end{aligned}
   \right.
\end{equation}
where  $\rho,\omega$ are the polar coordinates in the plane, $n\in\mathbb{Z}$, $z\in\mathbb{N}$
and $D=(\sqrt{2})^{-\max(k,n)}$. We choose $k=1$ and $n=-5$. 
The solution is given by $u(\rho,\omega)=-\frac{Dk}{n}\left(\frac{\rho}{R}\right)^{(n-k)}\rho^k\sin(n\omega)+R\log \rho$. The solution and the source terms are depicted in Figure~\ref{fig:pb1}.
Notice the variation in angular frequency as a function of the radius.

\subsubsection{Test problem 2: Four-lobe structure}

\begin{figure}[H]
    \centering
    \includegraphics[width=0.8\linewidth]{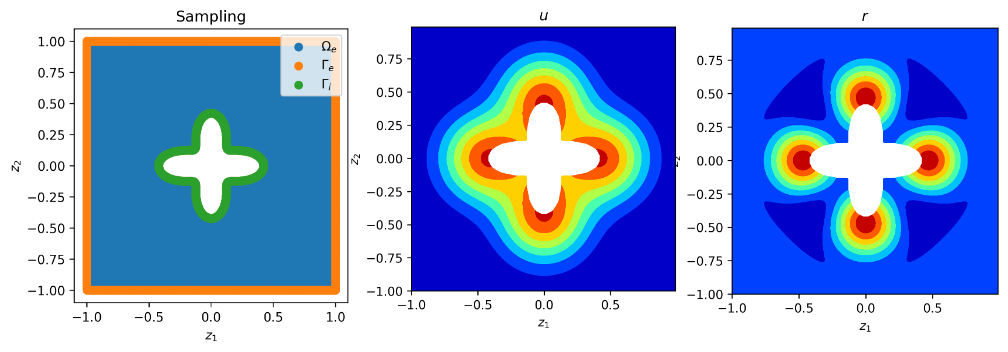}
    \caption{Test problem 2 (cf. \eqref{probleme_poisson},
      \eqref{pb2}). Left: the domain $\Omega$ and its
      subdomains. Center: the solution $u$. Right: the source term $r$.} 
    \label{fig:pb2}
\end{figure}

The domain for the second problem is the unit square
$\Omega = [0,1]^2$ with an embedded  surface defined by $\rho(\omega)
= R_m + R_d\cos(4\omega)$ with $R_m = 0.0305$, $R_d=0.117$. We
consider Dirichlet boundary conditions for both $\Gamma_e$ and
$\Gamma_i$.  The source term and boundary term are given by
\begin{equation}\label{pb2}
   \left\{
   \begin{aligned}
      &r(\rho,\omega) = 12(10\rho^2-1)e^{-10^2} + \sum_{k=1}^{4}40(10r_k^2-1) \text{ in }\Omega \\   
      &g(\rho,\omega) = 0.3e^{-10 \rho^2} + \sum_{k=1}^4e^{-10r_k^2}\text{ on }\Gamma_e \cup \Gamma_i\\
      &r_k = \sqrt{(x\pm0.45)^2+(y\pm0.45)^2}, \; k=1,\dots,4
   \end{aligned}
   \right.
\end{equation}
where $\rho$ and $\omega$ are the polar coordinates, and $x$ and $y$ the corresponding Cartesian coordinates. The solution is
given by $u(\rho,\omega)=0.3e^{-10 \rho^2} +
\sum_{k=1}^4e^{-10r_k^2}$. The solution and the source terms are
depicted in Figure \ref{fig:pb2}. 

\subsubsection{Test problem 3: Annulus domain with homogeneous source}

\begin{figure}[H]
    \centering
    \includegraphics[width=0.8\linewidth]{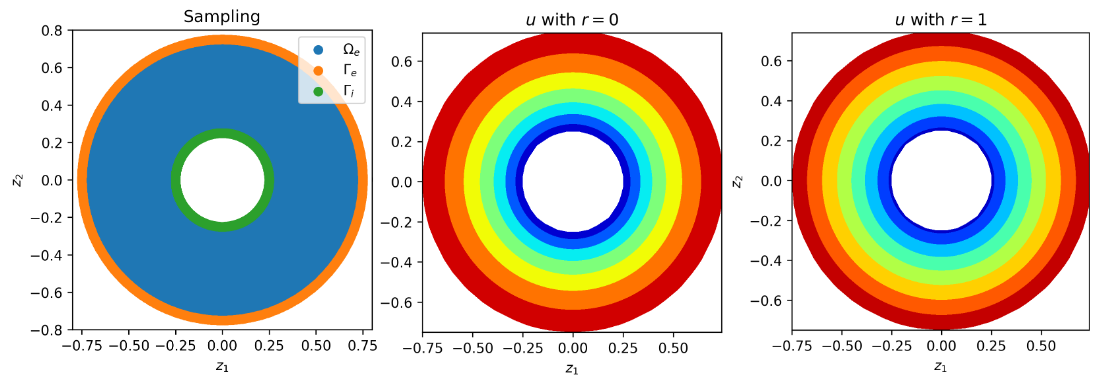}
    \caption{Test problems 3 (cf. \eqref{probleme_poisson}, \eqref{pb3}) and 4 (cf. \eqref{probleme_poisson}, \eqref{pb4}). Left: the domain $\Omega$ and its subdomains. Center: the solution $u$ for Test Problem 3. Right: the solution $u$ for Test Problem 4.}
    \label{fig:pb3}
\end{figure}

We now consider a domain defined by an annulus with inner radius
$R_i=0.25$ and outer radius $R_o=0.75$, and consider a Neumann
boundary condition on $\Gamma_i$ and a Dirichlet boundary condition on
$\Gamma_e$.  The source term and boundary term are given by  
\begin{equation}\label{pb3}
   \left\{
   \begin{aligned}
      &r(\rho,\omega) = 0 \text{ in }\Omega \\   
      &g(\rho,\omega) = 0\text{ on }\Gamma_e \\
      &h(\rho,\omega) = 1 \text{ on }\Gamma_i \\
   \end{aligned}
   \right.
\end{equation}

where $\rho$ and $\omega$ are the polar coordinates. The solution is
given by $u(\rho,\omega)=R_i\log(\frac{\rho}{R_0})$, which is depicted
in Figure \ref{fig:pb3}. 

\subsubsection{Test problem 4: Annulus domain with inhomogeneous source}

We finally consider again the domain defined by an annulus with inner
radius $R_i=0.25$ and the outer radius $R_o=0.75$ with Neumann
boundary condition on $\Gamma_i$ and Dirichlet boundary condition on
$\Gamma_e$.  The source and boundary terms now given by  
\begin{equation}\label{pb4}
   \left\{
   \begin{aligned}
      &r(\rho,\omega) = 1 \text{ in }\Omega \\   
      &g(\rho,\omega) = 0\text{ on }\Gamma_e \\
      &h(\rho,\omega) = 1 \text{ on }\Gamma_i.\\
   \end{aligned}
   \right.
\end{equation}
The solution is then given by $u(\rho,\omega)=\frac{\rho^2-R_0^2}{4}+R_i(1-\frac{R_i}{2})\log(\frac{\rho}{R_o})$.

\subsection{Network architectures}

We now give the details of the network architectures used in our second set of experiments.
It is important to notice that our method is not just applicable to
P-WWP networks, but can be used to train any architecture  composed of
sum of subnetworks.  

\begin{table}[H]
\centering
\begin{tabular}{|l|l|l|l|}
\hline
             & input scaling                      & First Activation & Other Activations \\ \hline
subnetwork 1 & None  & $\tanh$         & $\tanh$            \\ \hline
subnetwork 2 & $\mathcal{N}(0,20)$ & SFM(0.5)         & $\tanh$             \\ \hline
subnetwork 3 & $\mathcal{N}(0,40)$ & SFM(0.5)         & $\tanh$             \\ \hline
subnetwork 4 & $\mathcal{N}(0,60)$ & SFM(0.5)         & $\tanh$             \\ \hline
\end{tabular}
\end{table}
Each of the subnetworks is composed of 3 hidden layers of 100 neurons each. In order to assess computational performance, we also consider the standard single level training applied on the complete network.

\subsection{Training setup}

We will use the same setup as in Section~\ref{sec:PINNs} except for
the coefficients $\lambda_{\Omega_e}=1,\lambda_{\Gamma_e}=100$ and
$\lambda_{\Gamma_i}=1$ that weight internal and boundary losses.  

In order to compare the convergence speeds of the training methods, we
decided to consider the computational cost of the training, as the
epochs for the two training methods do not have the same cost.  For
this purpose, we consider as a unit of cost the price of optimising
our complete neural network over an epoch. When a subnetwork is
selected in the course of the \al{ML-BCD} algorithm, the unselected
part of the complete network is cached in memory and does not
contribute to the subnetwork optimization cost. 
Since the cost per iteration is linear in the number of parameters when using first-order methods, 
the cost of optimizing for an epoch one of four subnetworks of identical sizes is $\frac{1}{4} = 0.25$ cost units.
For these experiments we chose to do 1000 epochs in full network cycle
and 4000 epochs in sub-network cycle. Thus the cost of the full and
partial training is similar. We do a total of 9 cycles with 1000
epochs on the full network to start the training, for a total
computational cost of $1000 + 9\times(1000 + \frac{4000}{4}) = 19000$
units. For each problem we first compute the curve of the median
values over 10 runs of the loss and of the MSE, as a function of
epochs and for two different computational budgets (10k and 19k
units). We then select the lowest median loss obtained for a given
budget and record the associated median MSE.  

\subsection{Numerical results}

Table~\ref{tab:FAML-results} reports the results just described,
obtained with the Frequency-Aware-Multilevel (FAML). 

\begin{table}[H]
\centering

\begin{tabular}{|l|l|c|c|c|c|}
\hline
Problem   & Budget &   MSE    & MSE FAML &   Loss   & Loss FAML \\ \hline
Test pb 1 & 10000  & 1.40E-05 & 2.54E-07 & 1.64E-01 &  1.18E-02 \\ \cline{2-6}
          & 19000  & 7.70E-06 & 2.49E-07 & 6.33E-02 &  4.89E-03 \\ \hline
Test pb 2 & 10000  & 1.66E-05 & 3.99E-07 & 1.45E-01 &  1.17E-02 \\ \cline{2-6}
          & 19000  & 7.77E-06 & 2.57E-07 & 5.38E-02 &  4.42E-03 \\ \hline
Test pb 3 & 10000  & 1.05E-05 & 1.36E-07 & 1.04E-01 &  6.76E-03 \\ \cline{2-6}
          & 19000  & 3.81E-06 & 1.11E-07 & 3.71E-02 &  2.55E-03 \\ \hline
Test pb 4 & 10000  & 9.53E-06 & 1.62E-07 & 1.02E-01 &  7.40E-03 \\ \cline{2-6}
          & 19000  & 3.58E-06 & 1.32E-07 & 3.75E-02 &  2.63E-03 \\ \hline
\end{tabular}

\caption{Best median the loss and associated MSE for standard
  single-level and FAML training (10 independent
  runs)\label{tab:FAML-results}} 
\end{table}

\noindent
In each case, the proposed multi-level training yields lower losses
than the standard one. The differences are particularly significant
for a small computational budget where the improvement is of several
orders of magnitude. The multi-level training also always results in a
lower associated MSE. 

\begin{figure}[H]
    \centering
    \includegraphics[width=\linewidth]{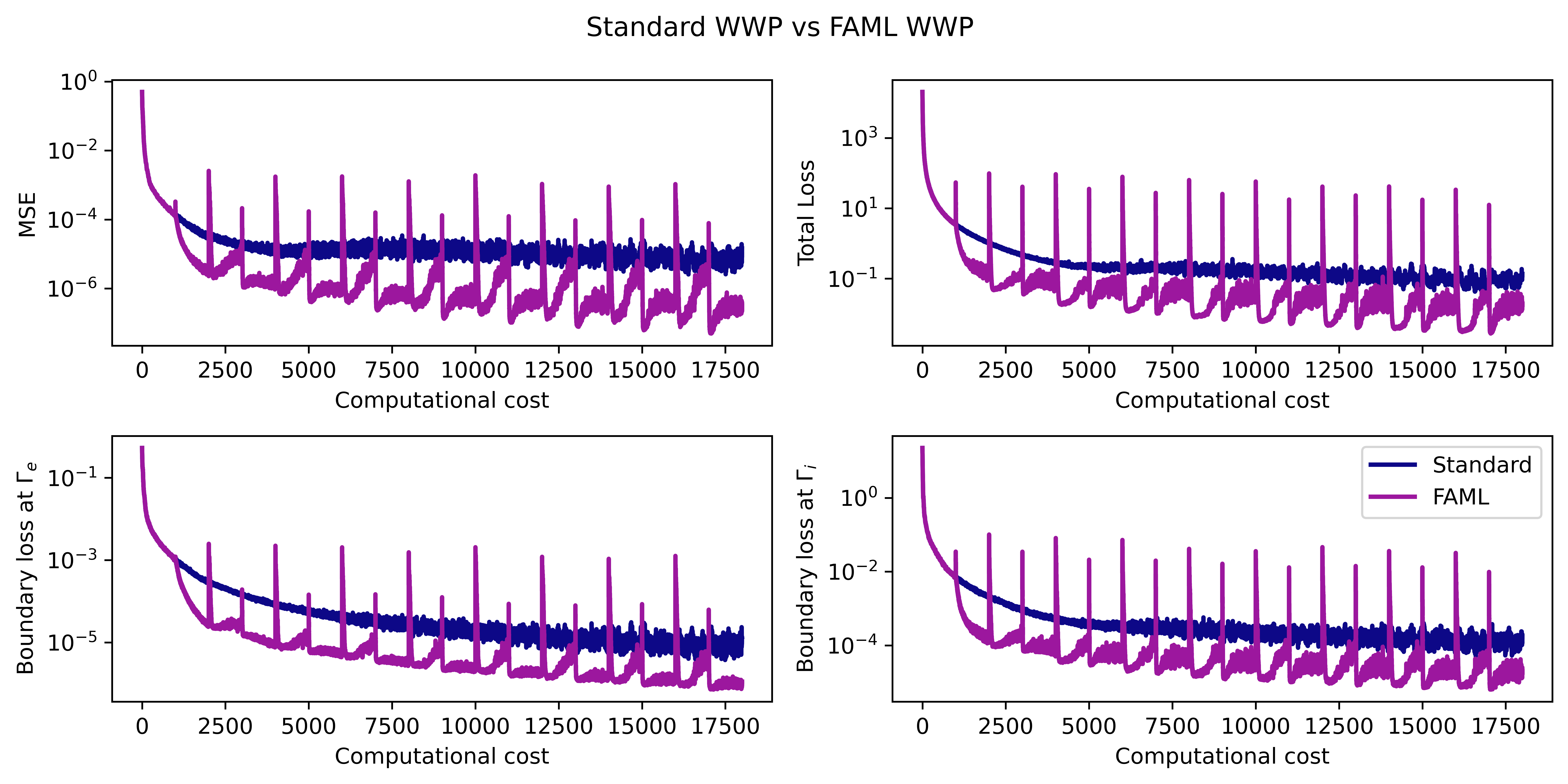}
    \label{pb3:wwp}
    \caption{Evolution of MSE and loss values for a typical run on Test problem 3}
\end{figure}

To provide further insight, we finally provide a typical example
allowing the comparison of standard and multi-level training: for Test
problem 3, we illustrate in  Figure \ref{pb3:wwp} the decrease of the
total loss and of its boundary and interior components as well as that
of the MSE as a function of computational cost. The multi-level
approach clearly outperforms the standard one. (We remind the reader
that peaks correspond to the restarts of the optimizer.) 

\section{Conclusion and perspectives}\label{sec:conclusion}

We first developed a new point of view on multilevel optimization
methods, arguing they can be seen as block-coordinate minimization
methods in a higher dimensional space. Distinguishing two contexts
(hierarchical and distributed),
we reformulated a class of multilevel algorithms in block form and
showed how convergence results for block-coordinate methods can be
applied to the multilevel case.  

We next illustrated this approach for the approximate solution of
partial differential equations, using physics-informed neural
networks as a block solver, and presented numerical
experiments with a method using pure alternating training (in the
hierarchical context) and a more elaborate frequency-aware technique
(in the distributed one) on complex Poisson problems. 
On these problems, both resulting multilevel PINNS methods
consistently produced lower losses than those obtained using
conventionally trained networks. Convergence was also shown to be much
faster allowing a considerable reduction of the computational cost to
obtain good solutions (often better than those obtained by a classical
training). 

While these initial results are encouraging, more research remains
desirable for a better understanding of the algorithms. Their
dependence on techniques to transfer algorithmic hyperparemeters
between levels is of particular interest. Applying our approach to
more general operator learning is also worth further investigation. 

\bibliographystyle{plain}
\bibliography{gmrt1v2}

\appendix


\section*{Appendix}
\section{Proof of Theorem~\ref{th:complexity}}\label{proof_complexity}

Consider applying the \al{ML-BCD} algorithm to the minimization of the
objective function $f$, which we assume is bounded below by
$f_{\rm low}$ and has a Lipschitz continuous gradient (with Lipschitz
constant $L$). Let $\alpha<1/L$ be the fixed stepsize used by the
algorithm.  Consider iteration $k$ and suppose that this iteration
occurs in the minimization of subproblem $\calA_i$ on the variables given by  $x_i$.
From the Lipschitz continuity of the gradient and \req{x-recur}, we obtain that
\[
f(x_{k+1})
\leq f(x_k) - \alpha \nabla_x f(x_k)^T \barg_{i,k} +\frac{\alpha^2L}{2}\|\barg_{i,k}\|^2
= f(x_k) - \alpha \|\barg_{i,k}\|^2+\frac{\alpha^2L}{2}\|\barg_{i,k}\|^2
<  f(x_k) - \frac{\alpha}{2} \|\barg_{i,k}\|^2.
\]
Since the minimization of subproblem $\calA_i$ has not yet terminated,
we must have that $\|\barg_{i,k}\|=\|g_{i,k}\|\geq \tau\epsilon$, and thus
\[
f(x_{k+1}) <  f(x_k) - \frac{\alpha\tau^2\epsilon^2}{2}.
\]
Summing this inequality on all iterations, we deduce that, for all $k$ before
termination of the \al{ML-BCD} algorithm,
\[ 
f(x_0)- f_{\rm low}
\geq f(x_0)-f(x_{k+1})
= \sum_{j=0}^k \Big(f(x_j)-f(x_{j+1})\Big)
\geq \frac{(k+1)\alpha\tau^2\epsilon^2}{2}.
\]
This implies that the algorithm cannot take more than
\[
\frac{2(f(x_0)- f_{\rm low})}{\alpha \tau^2 \epsilon^2}-1
\]
iterations before it teminates, proving the theorem with 
\[
\kappa_* =
\frac{2(f(x_0)- f_{\rm low})}{\alpha \tau^2}> \frac{2L(f(x_0)- f_{\rm low})}{\tau^2}.
\]
\epr

\section{A Mscale FAML variant and its performance}
\label{mscalefaml}
\subsection{Another frequency-aware architecture}

\begin{figure}[H]
    \centering
    \includegraphics[width=\linewidth]{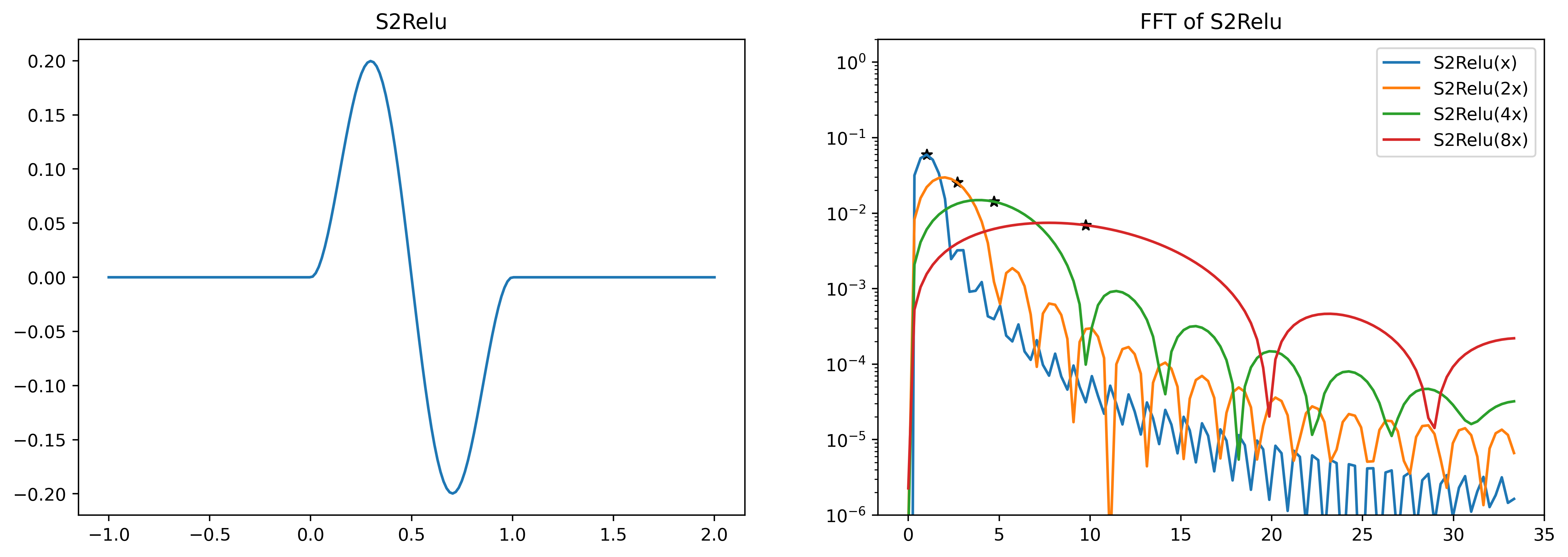}
    \caption{The s2Relu activation function (left) and its Fourier
      transform for several scalings (right). The black stars
      highlight the amplitude peaks.} 
    \label{fig:s2Relu}
\end{figure}

This appendix reports on tests conducted using a frequency-aware
network architecture based on MscaleDNN defined in
\cite{LiuCaiXu20}, instead of the WWP networks used in
Section~\ref{MPINNV2}. A large part of the success of Mscale networks
is due to their wavelet-inspired activation functions. These
compactly-supported have good scale separation properties and are
constructed so that the bandwidth of their Fourier transform
increases with their input scaling. 
Several activation functions have been proposed in
\cite{LiuCaiXu20}, but the most efficient one from a
practical point of view has been introduced in
\cite{Li20} and is given by 
\begin{equation}\label{s2relu-def}
\text{s2ReLU}(z) = \sin(2\pi z) \text{ReLU}(-(z-1)) \text{ReLU}(z),
\end{equation}
where $z$ is the input scaling.
This is a continuous function which decays faster and has better
localization property in the frequency domain than the previously
proposed sRelu. The amplitude peaks of the differently scaled s2ReLU
in the frequency domain are well separated in the Fourier domain. They
are indicated by black stars in Figure \ref{fig:s2Relu}, and we
observe their expected monotonic growth with scaling. 

As in Section \ref{MPINNV2}, we  use these functions to create a
neural network composed of a sum of subnetworks each targeting a
different frequency range (analogously to Figure~\ref{fig:pwwp}). In
this modified architecture, the lower frequency networks still use
$\tanh$ activation functions while the higher frequency networks now
use (\ref{s2relu-def}). For each subnetwork, the first activation
function is a $SFM$ function associated with a fixed input scaling, as
in standard Mscales networks. The details of this architecture are
given in Table \ref{table:MSCALE}. Each of the subnetworks is composed
of 3 hidden layers of 100 neurons each. 

\begin{table}[H]
\centering
\begin{tabular}{|l|l|l|l|}
\hline
             & input scaling      & First Activation & Other Activations \\ \hline
subnetwork 1 & (0.5,1,1.5,...,10) & SFM(1.0)         & $\tanh$             \\ \hline
subnetwork 2 & (11,12,...,29,30)  & SFM(0.5)         & s2ReLU           \\ \hline
subnetwork 3 & (31,32,...,50,51)  & SFM(0.5)         & s2ReLU           \\ \hline
subnetwork 4 & (51,52,...,69,70)  & SFM(0.5)         & s2ReLU           \\ \hline
\end{tabular}
\caption{Details of the parallel Mscale architecture\label{table:MSCALE}}
\end{table}

Associated with the \al{ML-BCD} algorithm (exactly as in
Section~\ref{MPINNV2}), this modified architecture defines an Mscale
variant of the FAML approach.  

\subsection{Numerical results}

We tested this approach using the same experimental setup and
methodology as that of Section~\ref{MPINNV2} and again compared its
performance to that the standard single level training applied on the
complete network.  The results are reported in
Table~\ref{tab:Mscale-res}. 

\begin{table}[H]
\centering
\begin{tabular}{|l|l|c|c|c|c|}
\hline
Problem   & Budget &   MSE    & MSE FAML &   Loss   & Loss FAML \\ \hline
Test Pb 1 & 10000  & 1.33E-05 & 2.76E-06 & 2.49E-01 &  7.98E-03 \\ \cline{2-6}
          & 19000  & 1.85E-06 & 2.51E-06 & 4.53E-03 &  2.50E-03 \\ \hline
Test Pb 2 & 10000  & 1.22E-05 & 3.87E-07 & 1.73E-01 &  8.14E-04 \\ \cline{2-6}
          & 19000  & 6.11E-07 & 2.74E-08 & 1.22E-05 &  3.87E-07 \\ \hline
Test Pb 3 & 10000  & 1.06E-05 & 1.85E-07 & 2.80E-01 &  1.06E-03 \\ \cline{2-6}
          & 19000  & 4.09E-07 & 1.23E-08 & 6.17E-04 &  2.09E-04 \\ \hline
Test Pb 4 & 10000  & 1.56E-05 & 1.98E-07 & 2.84E-01 &  1.04E-03 \\ \cline{2-6}
          & 19000  & 5.24E-07 & 1.21E-08 & 6.26E-04 &  1.85E-04 \\ \hline
\end{tabular}
\caption{Best median the loss and associated MSE for standard
  single-level and Mscale-FAML training (10 independent
  runs)\label{tab:Mscale-res}} 
\end{table}
As was the case when using P-WWP networks, the Mscale FAML variant
produced lower losses than the standard training in each case. The
differences are particularly significant for a small computational
budget where the improvement is of several orders of magnitude. This
is also almost always the case for the associated MSE. These results
thus remain excellent, despite the fact that our complexity theory
does not formally cover the Mscale FAML variant because of the
non-smoothness of (\ref{s2relu-def}).

\end{document}